# Deep Architectures for Automated Seizure Detection in Scalp EEGs

**Abstract**

Automated seizure detection using clinical electroencephalograms is a challenging machine learning problem because the multichannel signal often has an extremely low signal to noise ratio. Events of interest such as seizures are easily confused with signal artifacts (e.g, eye movements) or benign variants (e.g., slowing). Commercially available systems suffer from unacceptably high false alarm rates. Deep learning algorithms that employ high dimensional models have not previously been effective due to the lack of big data resources. In this paper, we use the TUH EEG Seizure Corpus to evaluate a variety of hybrid deep structures including Convolutional Neural Networks and Long Short-Term Memory Networks. We introduce a novel recurrent convolutional architecture that delivers 30% sensitivity at 7 false alarms per 24 hours. We have also evaluated our system on a held-out evaluation set based on the Duke University Seizure Corpus and demonstrate that performance trends are similar to the TUH EEG Seizure Corpus. This is a significant finding because the Duke corpus was collected with different instrumentation and at different hospitals. Our work shows that deep learning architectures that integrate spatial and temporal contexts are critical to achieving state of the art performance and will enable a new generation of clinically-acceptable technology.

## Introduction

Electroencephalograms (EEGs) are used in a wide range of clinical settings to record electrical activity along the scalp. Scalp EEGs are the primary means by which physicians diagnose brain-related illnesses such as epilepsy and seizures (Obeid and Picone 2017). However, manual analysis of EEG signals requires a highly trained board-certified neurophysiologist, and is a process that is known to have relatively low inter-rater agreement (IRA) (Swisher et al. 2015). It is a time-consuming and expensive process since the volume and velocity of the data far exceeds the available resources for detailed interpretation in real time. Automated analysis can improve the quality of patient care by reducing manual error and latency. In this paper, we focus on the specific problem of seizure detection, though the work presented here is also applicable to other EEG problems such as signal event detection (Harati et al. 2016) and abnormal detection (Lopez et al. 2015).

Like most machine learning problems of this nature, many algorithms have been applied including time–frequency analysis methods (Gotman et al. 1982) and nonlinear techniques (Schad et al. 2008). Despite much research progress, commercially available automated EEG analysis systems are impractical due to high false detection rates (Ramgopal 2014). Servicing false alarms in critical care settings places too much of a cognitive burden on caregivers, and hence, the outputs from these systems are ignored (Christensen et al. 2014). This creates quality of healthcare issues as well as cost and efficiency challenges.

Although contemporary approaches for automatic interpretation of EEGs have employed modern machine learning approaches (Alotaiby et al. 2014), deep learning algorithms that employ high dimensional models have not previously been utilized because there has been a lack of big data resources. A significant resource (Golmohammadi et al. 2017), known as the TUH EEG Seizure Corpus (TUSZ), has recently become available for EEG interpretation creating a unique opportunity to advance technology.

The goal of this work is to demonstrate that advanced deep learning approaches that have been successful in tasks like image processing and speech recognition, where ample amounts of annotated training data are available, can be applied to EEG interpretation. To achieve this goal, we evaluated several on a standard seizure detection task. We propose a novel deep learning architecture that reduces the false alarm rate while maintaining sensitivity and specificity. We demonstrate that the performance of this system is now approaching clinical acceptance.

## Exploiting Spatio-Temporal Context

Spatial and temporal context are required for accurate disambiguation of seizures from artifacts (Obeid and Picone 2017). In Figure 1, we show our generic architecture for processing EEG signals. The multichannel signal is sampled at 250 Hz using 16 bits of resolution, converted to a feature-based representation, processed through a sequential modeler, and then postprocessed using a variety of statistical models that impose constraints based on subject matter expertise. Several architectures that implement Gaussian Mixture Models (GMMs), hidden Markov model (HMMs) and deep learning (DL) have been evaluated.

Feature extraction, which is not the primary focus of this paper, typically relies on time frequency representations of the signal. Though we can replace traditional model-based feature extraction with deep learning-based approaches that operate directly on the sampled data, in this work we focus on the use of traditional cepstral-based features (Picone 1993). The use of more advanced discriminative features (Zhang et al. 2016) has not yet produced substantial improvements in performance for this application. Our system uses a standard linear frequency cepstral coefficient-based feature extraction approach (LFCCs) (Harati 2015; Lopez 2016). We also use first and second derivatives of the features since these improve performance.

Neurologists typically review EEGs in 10 sec windows and identify events with a resolution of approximately 1 sec. We analyze the signal in 1 sec epochs, and further divide this interval into 10 frames of 0.1 secs each so that features are computed every 0.1 seconds (referred to as the frame duration) using 0.2 second analysis windows (referred to as the window duration). The output of our feature extraction process is a feature vector of dimension 26 for each of 22 channels, with a frame duration of 0.1 secs.

## Sequential Decoding Using HMMs

HMMs are among the most powerful statistical modeling tools available today for signals that have both a time and frequency domain component (Picone 1990). HMMs have been used extensively in sequential decoding tasks like speech recognition to model the temporal evolution of the signal. Automated interpretation of EEGs is a problem like speech recognition since both time domain (e.g., spikes) and frequency domain information (e.g., alpha waves) are used to identify critical events (Obeid and Picone 2017).

In this study, a left-to-right channel-independent GMM-HMM, as illustrated in Figure 1, was used as a baseline system for sequential decoding. HMMs are attractive because training is much faster than comparable deep learning systems, and HMMs tend to work well when ample amounts of annotated data are available. We divide each channel of an EEG into 1-second epochs, and further subdivide these epochs into a sequence of frames. Each epoch is classified using an HMM trained on the subdivided epoch, and then these epoch-based decisions are postprocessed by additional statistical models in a process that parallels the language modeling component of a speech recognizer. Standard three state left-to-right HMMs (Picone 1990) with 8 Gaussian mixture components per state were used to model each channel of the 22-channel signal. A diagonal covariance matrix assumption was used for each mixture component. Channel-independent models were trained since channel-dependent models did not provide any improvement in performance.

Supervised training based on the Baum-Welch reestimation algorithm was used to train two models – seizure and background. Models were trained on segments of data containing seizures based on manual annotations that are available as part of TUSZ. Since seizures comprise a small percentage of the overall data (3% in the training set; 8% in the evaluation set), the amount of non-seizure data was limited to be comparable to the amount of seizure data, and non-seizure data was selected to include a rich variety of artifacts such as muscle and eye movements. Twenty iterations of Baum-Welch were used though performance is not very sensitive to this value. Standard Viterbi decoding (no beam search) was used in recognition to estimate the model likelihoods for every epoch of data (the entire file was not decoded as one stream because of the imbalance between the seizure and background classes – decoding was restarted for each epoch).

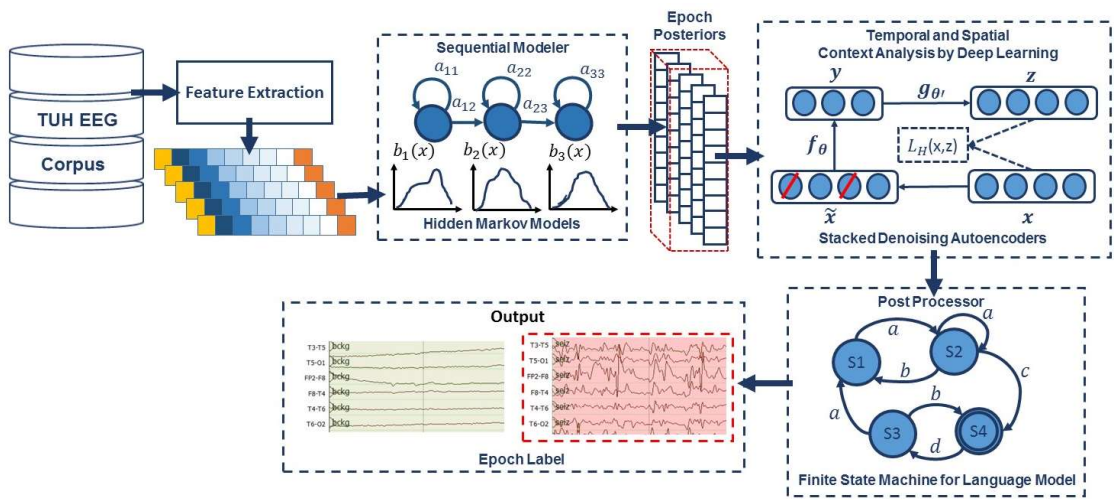

*Figure 1: A hybrid architecture for automatic interpretation of EEGs that integrates temporal and spatial context for sequential decoding of EEG events is shown. Two levels of postprocessing are used.*

The output of the epoch-based decisions was postprocessed by a deep learning system. Our baseline system used a Stacked denoising Autoencoder (SdA) (Vincent et al. 2008) as shown in Figure 1. SdAs are an extension of the stacked autoencoders and are a class of deep learning algorithms well-suited to learning knowledge representations that are organized hierarchically (Bengio et al. 2007). They also lend themselves to problems involving training data that is sparse, ambiguous or incomplete. Since inter-rater agreement is relatively low for seizure detection (Swisher et al. 2015), it made sense to evaluate this type of algorithm as part of a baseline approach.

An N-channel EEG was transformed into N independent feature streams using a standard sliding window based approach. The hypotheses generated by the HMMs were postprocessed using a second stage of processing that examines the temporal and spatial context. We apply a third pass of postprocessing that uses a stochastic language model to smooth hypotheses involving sequences of events so that we can suppress spurious outputs. This third stage of postprocessing provides a moderate reduction in false alarms.

Training of SdA networks are done in two steps: (1) pre-training in a greedy layer-wise approach (Bengio et al. 2007) and (2) fine-tuning by adding a logistic regression layer on top of the network (Hinton et al. 2006). The output of the first stage of processing is a vector of two likelihoods for each channel at each epoch. Therefore, if we have 22 channels, which is typical for an EEG collected using a standard 10/20 configuration (Obeid and Picone 2016), and 2 classes (seizure and background), we will have a vector of dimension 2 x 22 = 44 for each epoch.

Each of these scores is independent of the spatial context (other EEG channels) or temporal context (past or future epochs). To incorporate context, we form a supervector consisting of N epochs in time using a sliding window approach. We find benefit to making N large – typically 41. This results in a vector of dimension 1,804 that needs to be processed each epoch. The input dimensionality is too high considering the amount of manually labeled data available for training and the computational requirements. To deal with this problem we used Principal Components Analysis (PCA) (Ross et al. 2008) to reduce the dimensionality to 20 before applying the SdA postprocessing.

The parameters of the SdA model are optimized to minimize the average reconstruction error using a cross-entropy loss function. In the optimization process, a variant of stochastic gradient descent is used called "Minibatch stochastic gradient descent" (MSGD) (Zinkevich et al. 2010). MSGD works identically to stochastic gradient descent, except that we use more than one training example to make each estimate of the gradient. This technique reduces variance in the estimate of the gradient, and often makes better use of the hierarchical memory organization in modern computers.

The SdA network has three hidden layers with corruption levels of 0.3 for each layer. The number of nodes per layer are: first layer = 800, second layer = 500, third layer = 300. The parameters for pre-training are: learning rate = 0.5, number of epochs = 150, batch size = 300. The parameters for fine-tuning are: learning rate = 0.1, number of epochs = 300, batch size = 100. The overall result of the second stage is a probability vector of dimension two containing a likelihood that each label could have occurred in the epoch. A soft decision paradigm is used rather than a hard decision paradigm because this output is smoothed in the third stage of processing. A more detailed explanation about the third pass of processing is presented in (Harati et al. 2016).

**Context Modeling Using LSTMs**

To improve our ability to model context, a hybrid system composed of an HMM and a Long Short Term Memory (LSTM) network (Hochreiter et al. 1997) was implemented. These networks are a special kind of recurrent neural network (RNN) architecture that is capable of learning long-term dependencies, and can bridge time intervals exceeding 1,000 steps even for noisy incompressible input sequences. This is achieved by multiplicative gate units that learn to open and close access to the constant error flow.

Like the HMM/SdA hybrid approach previously described, the output of the first pass is a vector of dimension 2 × 22 × the window length. Therefore, we also use PCA before LSTM to reduce the dimensionality of the data to 20. For this study, we used a window length of 41 for LSTM, and this layer is composed of one hidden layer with 32 nodes. The output layer nodes in this LSTM level use a sigmoid function. The parameters of the models are optimized to minimize the error using a cross-entropy loss function. Adaptive Moment Estimation (Adam) is used (Kingma et al. 2015) in the optimization process.

To explore the potential of LSTMs to encode long-term dependencies, we designed another architecture, where Incremental Principal Components Analysis (IPCA) was used for dimensionality reduction (Ross et al. 2008; Levy et al. 2000). LSTM networks which operate directly on features spanning long periods of time need more memory efficient approaches. IPCA has constant memory complexity, on the order of the batch size, enabling use of a large dataset without loading the entire dataset into memory. IPCA builds a low-rank approximation for the input data using an amount of memory which is independent of the number of input data samples. It is still dependent on the input data features, but changing the batch size allows for control of memory usage.

The architecture of our IPCA/LSTM system is presented in Figure 2. In the IPCA/LSTM system, samples are converted to features by our standard feature extraction method previously described. Next, the features are delivered to an IPCA layer for spatial context analysis and dimensionality reduction. The output of IPCA is delivered to a one-layer LSTM for classification. The input to IPCA has a dimension that is a multiplication of the number of channels, the feature vector length, the number of features per seconds and window duration (in seconds). We typically use a 7-second window duration, so the IPCA input is a vector of dimension $22 \times 26 \times 7 \times 10 = 4004$. A batch size of 50 is used in IPCA and the output dimension is 25. In order to learn long-term dependencies, one LSTM with a hidden layer size of 128 and batch size of 128 is used along with Adam optimization and a cross–entropy loss function.

## Two-Dimensional Decoding Using CNNs

Convolutional Neural Networks (CNNs) have delivered state of the art performance on highly challenging tasks such as speech (Saon et al. 2016) and image recognition (Simonyan et al. 2014). CNNs are usually comprised of convolutional layers along with subsampling layers which are followed by one or more fully connected layers. In the case of two dimensional CNNs that are common in image and speech recognition, the input to a convolutional layer is $W \times H \times N$ data (e.g. an image) where W and H are the width and height of the input data, and N is the number of channels (e.g. in an RGB image, $N = 3$). The convolutional layer will have K filters (or kernels) of size $M \times N \times Q$ where M and N are smaller than the dimension of the data and Q is typically smaller than the number of channels. In this way CNNs have a large learning capacity that can be controlled by varying their depth and breadth to produce K feature maps of size $(W-M+1) \times (H-N+1)$. Each map is then subsampled with max pooling over $P \times P$ contiguous regions. An additive nonlinearity is applied to each feature map either before or after the subsampling layer.

Our overall architecture of a system that combines CNN and a multi-layer perceptron (MLP) (Simonyan et al. 2014) is shown in Figure 3. The network contains six convolutional layers, three max pooling layers and two fully-connected layers. A rectified linear unit (ReLU) non-linearity is applied to the output of every convolutional and fully-connected layer (Nair et al. 2010). Drawing on an image classification analogy, each image is a signal where the width of the image (W) is the window length multiplied by the number of samples per second, the height of the image (H) is the number of EEG channels and the number of image channels (N) is the length of the feature vector.

In our optimized system, a window duration of 7 seconds is used. The first convolutional layer filters the input of size of $70 \times 22 \times 26$ using 16 kernels of size $3 \times 3$ with a stride of 1. The second convolutional layer filters its input using 16 kernels of size $3 \times 3$ with a stride of 1. The first max pooling layer takes as input the output of the second convolutional layer and applies a pooling size of $2 \times 2$. This process is repeated two times more with 32 and 64 kernels. Next, a fully-connected layer with 512 neurons is applied and the output is fed to a 2-way sigmoid function which produces a two-class decision (the final epoch label).

## Recurrent Convolutional Neural Networks

In our final architecture, which is shown in Figure 4, we integrate 2D CNNs, 1-D CNNs and LSTM networks, which we refer to as a CNN/LSTM, to better exploit long-term dependencies. Note that the way that we handle data in CNN/LSTM is different from the CNN/MLP system presented in Figure 3. Drawing on a video classification analogy, input data is composed of frames distributed in time where each frame is an image of width (W) equal to the length of a feature vector, the height (H) equals the number of EEG channels, and the number of image channels (N) equals one. Then input data consists of T frames where T is equal to the window length multiplied by the number of samples per second. In our optimized system with a window duration of 21 seconds, the first 2D convolutional layer filters 210 frames (T = 21 × 10) of EEGs distributed in time with a size of 26 × 22 × 1 (W = 26, H = 22, N = 1) using 16 kernels of size 3 × 3 and with a stride of 1. The first 2D max pooling layer takes as input a vector which is 260 frames distributed in time with a size of 26 × 22 × 16 and applies a pooling size of 2 × 2. This process is repeated two times with two 2D convolutional layers with 32 and 64 kernels of size 3 × 3 respectively and two 2D max pooling layers with a pooling size 2 × 2.

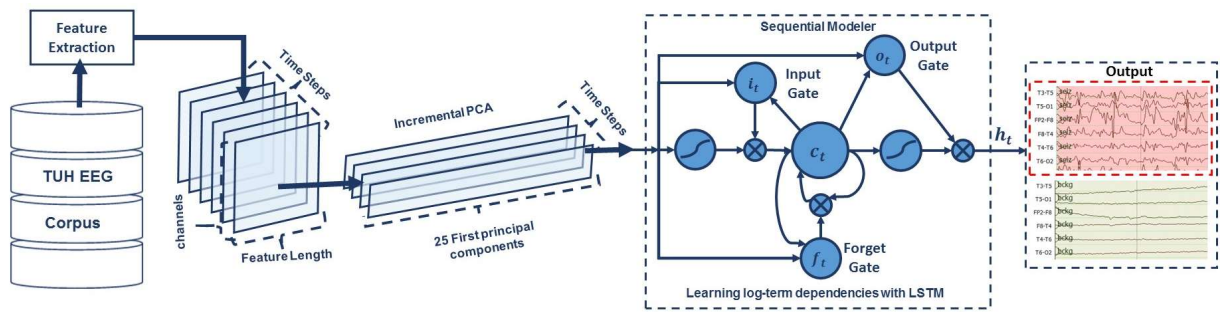

*Figure 2: An architecture that integrates IPCA for spatial context analysis and LSTM for learning long-term temporal dependencies*

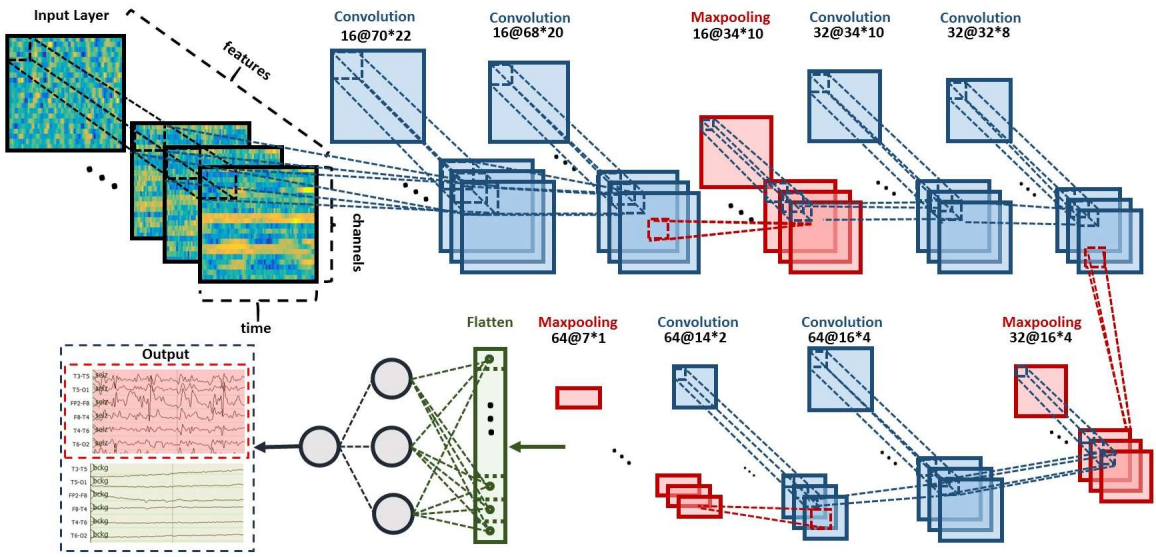

*Figure 3: Two-dimensional decoding of EEG signals using a CNN/MLP hybrid architecture is shown that consists of six convolutional layers, three max pooling layers and two fully-connected layers.*

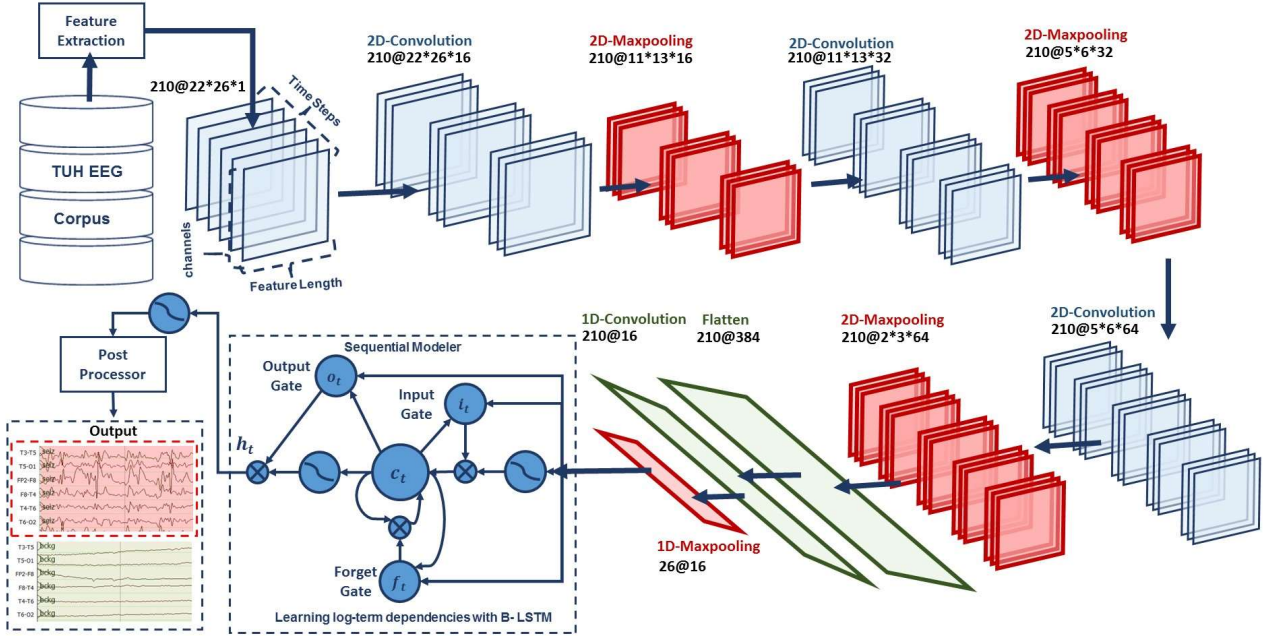

*Figure 4: A deep recurrent convolutional architecture for two-dimensional decoding of EEG signals that integrates 2D CNNs, 1-D CNNs and LSTM networks is shown.*

The output of third max pooling is flattened to 210 frames with size of 384 × 1. Then a 1D convolutional layer filters the output of the flattening layer using 16 kernels of size 3 which decreases the dimensionality in space to 210 × 16. Then we apply a 1D maxpooling layer with a size of 8 to decrease the dimensionality to 26 × 16. This is the input to a deep bidirectional LSTM network where the dimensionality of the output space is 128 and 256. The output of the last bidirectional LSTM layer is fed to a 2-way sigmoid function which produces a final classification of an epoch. To overcome the problem of overfitting and force the system to learn more robust features, dropout and Gaussian noise layers are used between layers (Srivastava et al. 2010). To increase non-linearity, Exponential Linear Units (ELU) are used (Clevert et al. 2017). Adam is used in the optimization process along with a mean squared error loss function.

## Experiments

The lack of big data resources that can be used to train sophisticated statistical models compounds a major problem in automatic seizure detection. Inter-rater agreement for this task is low, especially when considering short seizure events (Obeid and Picone 2017). Manual annotation of a large amount of data by a team of certified neurologists is extremely expensive and time consuming. In this study, we are reporting results for the first time on the TUSZ and a comparable corpus, DUSZ, from Duke University (Swisher et al, 2015). TUSZ was used as the training and test set corpus, while DUSZ was used as a held-out evaluation set. It is important to note that TUSZ was collected using several generations of Natus EEG equipment, while DUSZ was collected using Nihon Kohden equipment. Hence, this is a true open-set evaluation since the data were collected under completely different recording conditions. A summary of these corpora is shown in Table 1.

A comparison of the performance of the different architectures presented in this paper, for sensitivity in range of 30%, are shown in Table 2. The related DET curve is illustrated in Figure 5. These systems were evaluated using a method of scoring popular in the EEG research community known as the overlap method (Wilson et al. 2003). True positives (TP) are defined as the number of epochs identified as a seizure in the reference annotations and correctly labeled as a seizure by the system. True negatives (TN) are defined as the number of epochs correctly identified as non-seizures. False positives (FP) are defined as the number of epochs incorrectly labeled as seizure while false negatives (FN) are defined as the number of epochs incorrectly labeled as non-seizure. Sensitivity shown in Table 2 is computed as TP/(TP+FN). Specificity is computed as TN/(TN+FP). The false alarm rate is the number of FPs per 24 hours.

Table 1: An overview of the TUHS and Duke corpora

| Description | TUHS Train | TUHS Eval | Duke Eval |
|---|---|---|---|
| Patients | 64 | 50 | 45 |
| Sessions | 281 | 229 | 45 |
| Files | 1,028 | 985 | 45 |
| Seizure (secs) | 17,686 | 45,649 | 48,567 |
| Non-Seizure (secs) | 596,696 | 556,033 | 599,381 |
| Total (secs) | 614,382 | 601,682 | 647,948 |

Table 2: Performance on the TUSZ

| System | Sensitivity | Specificity | FA/24 Hrs. |
|---|---|---|---|
| HMM | 30.32% | 80.07% | 244 |
| HMM/SdA | 35.35% | 73.35% | 77 |
| HMM/LSTM | 30.05% | 80.53% | 60 |
| IPCA/LSTM | 32.97% | 77.57% | 73 |
| CNN/MLP | 39.09% | 76.84% | 77 |
| CNN/LSTM | 30.83% | 96.86% | 7 |

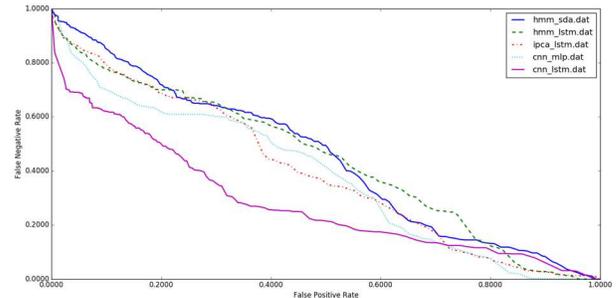

Figure 5: A DET curve comparing performance on TUSZ

It is important to note that the results are much lower than what is often published in the literature on other seizure detection tasks. This is due to a combination of factors including (1) the neuroscience community has favored a more permissive method of scoring that tends to produce much higher sensitivities and lower false alarm rates; and (2) TUSZ is a much more difficult task than any corpus previously released as open source. The evaluation set was designed to be representative of common clinical issues and includes many challenging examples of seizures.

Also, note that the HMM baseline system, which is shown in the first row of Table 2, operates on each channel independently. The other methods consider all channels simultaneously by using a supervector that is a concatenation of the feature vectors for all channels. The baseline HMM system only classifies epochs (1 sec in duration) using data from within that epoch. It does not look across channels or across multiple epochs when performing epoch-level classification. The results of the hybrid HMM and deep learning structures show that adding a deep learning structure for temporal and spatial analysis of EEGs can decrease the false alarm rate dramatically. Further, by comparing the results of HMM/SdA with HMM/LSTM, we find that a simple one layer LSTM performs better than 3 layers of SdA due to LSTM's ability to explicitly model long-term dependencies. Note that in this case the complexity and training time of these two systems is comparable.

The best overall system is the combination of CNN and LSTM. This doubly deep recurrent convolutional structure models both spatial relationships (e.g., cross-channel dependencies) and temporal dynamics (e.g., spikes). For example, CNN/LSTM does a much better job rejecting artifacts that are easily confused with spikes because these appear on only a few channels, and hence can be filtered based on correlations between channels. The depth of the convolutional network is important since the top convolutional layers tend to learn generic features while the deeper layers learn dataset specific features. Performance degrades if a single convolutional layer is removed. For example, removing any of the middle convolutional layers results in a loss of about 4% in the sensitivity.

We have also conducted an evaluation of our CNN/LSTM system on a DUSZ. The results are shown in Table 3. A DET curve is shown in Figure 5. At high false positive rates, performance between the two systems is comparable. At low false positive rates, false positives on TUSZ are lower than on DUSZ. This suggests there is room for additional optimizations on DUSZ.

In these experiments, we observed that the choice of optimization method had a considerable impact on performance. The results of our best performing system, CNN/LSTM, was evaluated using a variety of optimization methods, including SGD (Wilson et al. 2003), RMSprop (Bottou et al. 2004), Adagrad (Tieleman et al. 2012), Adadelta (Duchi et al. 2011), Adam (Kingma et al. 2015), Adamax (Kingma et al. 2015) and Nadam (Zeiler et al. 2013) as shown in Table 4. The best performance is achieved with Adam, a learning rate of $\alpha = 0.0005$, a learning rate decay of 0.0001, exponential decay rates of $\beta_1 = 0.9$ and $\beta_2 = 0.999$ for the moment estimates and a fuzz factor of $\epsilon = 10^{-8}$. The parameters follow the notation described in (Kingma et al. 2015). Table 4 also illustrates that Nadam delivers comparable performance to Adam. Adam combines the advantages of AdaGrad which works well with sparse gradients, and RMSProp which works well in non-stationary settings.

Similarly, we evaluated our CNN/LSTM using different activation functions, as shown in Table 5. ELU delivers a small but measurable increase in sensitivity, and more importantly, a reduction in false alarms. ReLUs and ELUs accelerate learning by decreasing the gap between the normal gradient and the unit natural gradient (Clevert et al. 2017). ELUs push the mean towards zero but with a significantly smaller computations footprint. But unlike ReLUs, ELUs have a clear saturation plateau in its negative regime, allowing them to learn a more robust and stable representation, and making it easier to model dependencies between ELUs.

## Conclusions

In this paper, we introduced a variety of deep learning architectures for automatic classification of EEGs including a hybrid architecture that integrates CNN and LSTM. While this architecture delivers better performance than other deep structures, its performance still does not meet the needs of clinicians. Human performance on similar tasks is in the range of 65% sensitivity with a false alarm rate of 12 per 24 hours (Swisher et al. 2015). The false alarm rate is particularly important to critical care applications since it impacts the workload experienced by healthcare providers.

The primary error modalities observed were false alarms generated during brief delta range slowing patterns such as intermittent rhythmic delta activity. A variety of these types of artifacts have been observed mostly during inter-ictal and post-ictal stages. Training models on such events with diverse morphologies has the potential to significantly reduce the remaining false alarms. This is one reason we are continuing our efforts to annotate a larger portion of TUSZ.

Table 3: Performance of CNN/LSTM on DUSZ

| Corpus | Sensitivity | Specificity | FA/24 Hrs. |
|--------|-------------|-------------|------------|
| TUSZ   | 30.83%      | 96.86%      | 7          |
| DUSZ   | 33.71%      | 70.72%      | 40         |

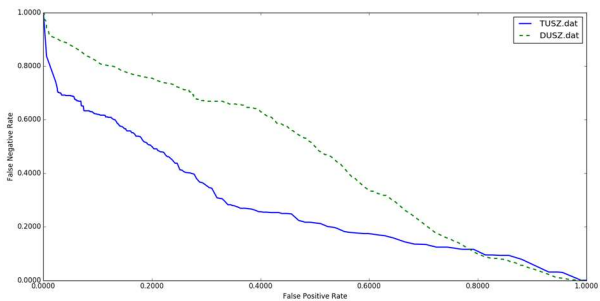

*Figure 6: A performance comparison of TUSZ and DUSZ*

Table 4: Comparison of optimization algorithms

| Opt.     | Sensitivity | Specificity | FA/24 Hrs. |
|----------|-------------|-------------|------------|
| SGD      | 23.12%      | 72.24%      | 44         |
| RMSprop  | 25.17%      | 83.39%      | 23         |
| Adagrad  | 26.42%      | 80.42%      | 31         |
| Adadelta | 26.11%      | 79.14%      | 33         |
| Adam     | 30.83%      | 96.86%      | 7          |
| Adamax   | 29.25%      | 89.64%      | 18         |
| Nadam    | 30.27%      | 92.17%      | 14         |

Table 5: Comparison of activation functions

| Activation | Sensitivity | Specificity | FA/24 Hrs. |
|------------|-------------|-------------|------------|
| Linear     | 26.46%      | 88.48%      | 25         |
| Tanh       | 26.53%      | 89.17%      | 21         |
| Sigmoid    | 28.63%      | 90.08%      | 19         |
| Softsign   | 30.05%      | 90.51%      | 18         |
| ReLU       | 30.51%      | 94.74%      | 11         |
| ELU        | 30.83%      | 96.86%      | 7          |